\pdfoutput=1

\documentclass[pdflatex,sn-mathphys-num]{sn-jnl}

\usepackage{graphicx}%
\usepackage{multirow}%
\usepackage{amsmath,amssymb,amsfonts}%
\usepackage{amsthm}%
\usepackage{mathrsfs}%
\usepackage[title]{appendix}%
\usepackage{xcolor}%
\usepackage{textcomp}%
\usepackage{manyfoot}%
\usepackage{booktabs}%
\usepackage{algorithm}%
\usepackage{algorithmicx}%
\usepackage{algpseudocode}%
\usepackage{listings}%
\usepackage{afterpage}  
\usepackage{longtable}
\usepackage{fancyhdr}

\usepackage[most]{tcolorbox}

\usepackage{cleveref}
\usepackage{caption}
\usepackage{subcaption}
\usepackage{enumitem}
\usepackage{multicol}
\usepackage{bm}
\usepackage{makecell}
\usepackage{mdframed} 
\usepackage{colortbl}
\usepackage{float}

\definecolor{my_gray}{rgb}{0.9216, 0.9216, 0.9216}
\definecolor{my_green}{rgb}{0.078, 0.686, 0.384}

\begin{document}

\pagestyle{fancy}
\fancyhf{} 
\fancyhead[R]{\thepage} 
\renewcommand{\headrulewidth}{0pt}
\makeatletter
\let\ps@plain\ps@fancy
\makeatother

\title[SciHorizon: Benchmarking AI-for-Science Readiness from Scientific Data to Large Language Models]{SciHorizon: Benchmarking AI-for-Science Readiness from Scientific Data to Large Language Models}

\author[1,2]{\fnm{Chuan} \sur{Qin}$^{*}$}
\author[1,2]{\fnm{Xin} \sur{Chen}$^{\dagger}$}
\author[1]{\fnm{Chengrui} \sur{Wang}}
\author[1]{\fnm{Pengmin} \sur{Wu}}
\author[2,3]{\fnm{Xi} \sur{Chen}}
\author[1]{\fnm{Yihang} \sur{Cheng}}
\author[1]{\fnm{Jingyi} \sur{Zhao}}
\author[1]{\fnm{Meng} \sur{Xiao}}
\author[1]{\fnm{Xiangchao} \sur{Dong}}
\author[1]{\fnm{Qingqing} \sur{Long}}
\author[1]{\fnm{Boya} \sur{Pan}}
\author[4]{\fnm{Han} \sur{Wu}}
\author[1,2]{\fnm{Chengzan} \sur{Li}}
\author[1,2]{\fnm{Yuanchun} \sur{Zhou}$^{\ddagger}$}
\author[5,6]{\fnm{Hui} \sur{Xiong}}
\author[1,2]{\fnm{Hengshu} \sur{Zhu}$^{\ddagger}$}

\affil[1]{Computer Network Information Center, Chinese Academy of Sciences}

\affil[2]{University of Chinese Academy of Sciences}

\affil[3]{University of Science and Technology of China}
\affil[4]{Hefei University of Technology}
\affil[5]{The Hong Kong University of Science and Technology (Guangzhou)}
\affil[6]{The Hong Kong University of Science and Technology}

\affil[]{Concat Email: \url{scihorizon@cnic.cn}}
\affil[]{Project Homepage: \url{www.scihorizon.cn}}

\renewcommand{\thefootnote}{\fnsymbol{footnote}}
\footnotetext[1]{Leading the LLM assessment.}
\footnotetext[2]{Leading the scientific data assessment.}
\footnotetext[3]{Project Leader.}
\renewcommand{\thefootnote}{\arabic{footnote}}

\maketitle
\begin{abstract}

In recent years, the rapid advancement of Artificial Intelligence (AI)  technologies, particularly Large Language Models (LLMs), has revolutionized the paradigm of scientific discovery, establishing AI-for-Science (AI4Science) as a dynamic and evolving field. However, there is still a lack of an effective framework for the overall assessment of AI4Science, particularly from a holistic perspective on data quality and model capability. Therefore, in this study, we propose SciHorizon, a comprehensive assessment framework designed to benchmark the readiness of AI4Science from both scientific data and LLM perspectives. First, we introduce a generalizable framework for assessing AI-ready scientific data, encompassing four key dimensions—Quality, FAIRness, Explainability, and Compliance—which are subdivided into 15 sub-dimensions. Drawing on data resource papers published between 2018 and 2023 in peer-reviewed journals, we present recommendation lists of AI-ready datasets for Earth, Life, and Materials Sciences, making a novel and original contribution to the field. Concurrently, to assess the capabilities of LLMs across multiple scientific disciplines, we establish 16 assessment dimensions based on five core indicators—Knowledge, Understanding, Reasoning, Multimodality, and Values—spanning Mathematics, Physics, Chemistry, Life Sciences, and Earth and Space Sciences. Using the developed benchmark datasets, we have conducted a comprehensive evaluation of over 50 representative open-source and closed-source LLMs. All the results are publicly available and can be accessed online at~\underline{\url{www.scihorizon.cn/en}}.

\end{abstract}

\section{Introduction}

Scientific data resources serve as the core driver propelling scientific discoveries and as a critical enabler for interdisciplinary research that integrates Artificial Intelligence (AI) with other disciplines. As AI technologies advance rapidly, particularly in Large Language Models (LLMs), AI-for-Science (AI4Science) has emerged as a ‌frontier research hotspot~\cite{zhang2019aftershock,zhu2020rapid,zhang2021exploiting,shen2019machine,hu2025scalable,ji2025comprehensive,wang2025segquadtree}.
For instance, in structural biology, DeepMind’s AlphaFold has revolutionized protein structure prediction by achieving unprecedented accuracy, effectively resolving a challenge that previously demanded decades of experimental effort~\cite{jumper2021highly}. Likewise, LLMs, such as OpenAI’s ChatGPT, are expanding AI’s role in scientific research by not only enhancing data analysis capabilities~\cite{hassani2023role,morgan2023exploring,qin2025cotr,gong2024graph}, but also assisting in literature review~\cite{temsah2023overview}, hypothesis generation~\cite{zhou2024hypothesis,park2024can}, and complex reasoning~\cite{frieder2023mathematical,guo2023can}. These models facilitate the synthesis of vast scientific knowledge, accelerate discovery processes, and foster interdisciplinary collaboration, thereby reshaping the landscape of modern scientific inquiry.

Despite rapid advancements in AI4Science, the field faces persistent challenges stemming from the reliance on large-scale, interdisciplinary datasets and the scarcity of AI-ready, high-quality data‌. Therefore, researchers advocate for systematic assessment frameworks that integrate metrics for data accuracy, completeness, and domain relevance.
While traditional tools are instrumental in assessing data quality from different perspectives~\cite{schelter2018automating,houston2018assessing,shi2019data,li2021cleanml} or emphasizing specific aspects of data readiness~\cite{clarke2019fairshake,holland2020dataset,arca2020entropy,carlini2022privacy}, they fall short of addressing the unique requirements of AI applications in the scientific domain. This leads to two challenges in evaluating scientific data in AI4Science: 1) AI researchers struggle to efficiently extract valuable insights from vast, domain-specific datasets, leading to the underutilization or misapplication of high-potential data that could drive significant advances in AI-driven scientific discovery; and 2) Researchers across disciplines lack clear criteria to assess whether their datasets align with AI model requirements, hindering the development and optimization of high-quality, AI-ready data.


Meanwhile, after assessing the possibility of high-quality scientific data integration, a critical challenge lies in comprehensively and effectively assessing the application capabilities of AI4Science models at a fine-grained level. Despite the proliferation of diverse assessment frameworks for LLMs~\cite{chang2024survey}, those specifically designed for scientific applications remain scarce. Recently, several targeted benchmarks have been proposed, however, they exhibit notable limitations: 1) Most of them focus on specific disciplines~\cite{cobbe2021training,singhal2023large}, lacking a unified framework that accommodates multiple scientific fields; 2) Their capability assessments are relatively narrow—for instance, JEEBench~\cite{arora-etal-2023-llms} primarily evaluates reasoning abilities related to basic computations, while MultiMedQA focuses on assessing clinical knowledge~\cite{singhal2023large}, both of which fall short of providing a more comprehensive and fine-grained evaluation; and 3) No existing framework assesses whether LLMs embody the correct scientific research values~\cite{wang2023scibench}, which is crucial for the responsible adoption of AI techniques in scientific tasks. 

To bridge these critical gaps, here we present SciHorizon—an integrated assessment framework that evaluates AI4Science readiness from both AI-ready data and LLM perspectives. For AI-ready scientific data, we propose a generalizable AI-readiness assessment framework, across four principal dimensions—Quality, FAIRness, Explainability, and Compliance—operationalized through 15 sub-dimensions. To demonstrate its applicability, we analyze approximately 1,600 datasets published between 2018 and 2023, primarily consisting of those published in data resource papers in peer-reviewed journals (e.g., \textit{Scientific Data} and \textit{ESSD}), identifying dataset recommendations for Earth, Life, and Materials Sciences to support AI-driven scientific advancements. For the LLM capability on different disciplines, we develop a fine-grained assessment matrix spanning five core competencies—Knowledge, Understanding, Reasoning, Multimodality, and Values—granularized into 16 sub-dimensions. Our benchmark suite, covering Mathematics, Physics, Chemistry, Life Sciences, and Earth and Space Sciences, enables systematic comparison of more than 50 representative open-source and closed-source LLMs. All the results are publicly available and can be accessed online at~\underline{\url{www.scihorizon.cn/en}}~\footnote{The platform was initially launched on January 22, 2025~\cite{scihorizon2025scihorizon}.}.

\section{Related Works}
The related works of this paper can be grouped into two categories, namely Data AI-Readiness and LLM Benchmarking.

\subsection{Data AI-Readiness}
The assessment of data AI-readiness has drawn more and more attention. Hiniduma et al. proposed a taxonomy of data readiness for AI metrics~\cite{ai360degree} and developed a quantitative assessment framework for data AI readiness~\cite{AIDRIN}. They are mainly from a data modality perspective and lack support and applicability for scientific data. Regarding the scientific data, the FAIR principles (Findable, Accessible, Interoperable, Reusable) presented in 2016 have been widely recognized as basic characteristics for scientific data sharing~\cite{FAIR}. The principles are also used for AI-readiness assessment for scientific data~\cite{FAIR4AI,AI-Fairness}. Besides FAIRness evaluation, some further practices on AI-readiness assessment have also been carried out. ESIP Data Readiness Cluster published a checklist to examine AI-readiness for open environmental datasets~\cite{ESIP}, covering data quality, data documentation, data access, and data preparation. The checklist serves more as a guideline than an evaluation and has less involvement in the AI applications. NIH Bridge to Artificial Intelligence (Bridge2AI) Standards Working Group presented criteria and methods for assessing the AI-readiness of biomedical data~\cite{bridge2AI}, which is a good practice for a specific discipline. Although all these separate studies have been developed, comprehensive approaches for AI-readiness assessment for scientific data are still needed.

\subsection{LLM Benchmarking}
With the rapid advancement of LLMs~\cite{wu2024survey,tong2025missteps,jiang2024enhancing}, numerous benchmarks have emerged to assess their performance across scientific disciplines. Early evaluations focused primarily on mathematical reasoning, such as GSM8K \cite{cobbe2021training} and MATH \cite{hendrycks2021measuring}. 
Recently, specialized benchmarks such as TheoremQA \cite{chen-etal-2023-theoremqa} and MathVista \cite{lu2024mathvista} have been introduced to assess theorem-based reasoning and multimodal mathematical understanding.
In addition to mathematics, researchers have developed broader scientific benchmarks to evaluate LLMs across multiple domains, such as ScienceQA \cite{lu2022learn}, JEEBench \cite{arora-etal-2023-llms}, SciEval \cite{sun2024scieval}, and BIG-Bench \cite{srivastava2023beyond}.
Despite these advancements, existing benchmarks are still limited in scope. Most focus on isolated disciplines~\cite{frieder2024mathematical}, without providing a unified framework for cross-disciplinary evaluation. 
Furthermore, many evaluations prioritize factual recall \cite{wan-etal-2024-logicasker, zhang2024sciglm}, overlooking the need for fine-grained assessments of scientific problem-solving and advanced reasoning. More critically, no benchmark systematically evaluates whether LLMs adhere to fundamental scientific research values—such as academic integrity, fairness, and transparency—which are crucial for their responsible use in scientific workflows \cite{wang2023scibench}.

\section{SciHorizon Framework}

\begin{figure*}
    \centering
    \includegraphics[width=\linewidth]{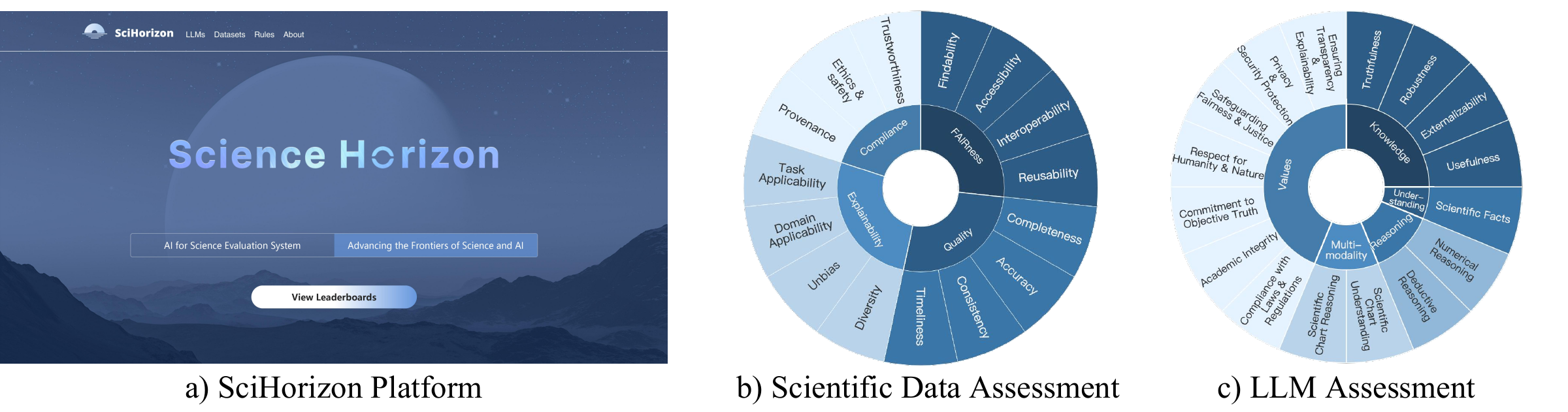}
    \caption{Overview of the SciHorizon platform.}
    \label{fig:fig1}
\end{figure*}


The framework of SciHorizon consists of two components: scientific data assessment and LLM assessment. Figure \ref{fig:fig1} presents an overview of the SciHorizon platform and its fine-grained assessment dimensions for both scientific data and LLMs.

\subsection{Scientific Data Assessment}
To systematically assess the AI readiness of scientific datasets, we have developed a comprehensive evaluation framework based on four key dimensions: \textbf{Quality}, \textbf{FAIRness}, \textbf{Explainability}, and \textbf{Compliance}. The following sections provide a detailed overview of this assessment framework and its practices.

\subsubsection{Quality}
This metric measures the Completeness, Accuracy, Consistency, and Timeliness of datasets to make sure the datasets are of high quality. We expand the traditional quality metrics by combining characteristics of scientific data.
\begin{itemize}[leftmargin=12pt]
    \item \textbf{Completeness} refers to the thorough recording of data elements and the inclusion of all necessary documents. This encompasses not only the data entity itself but also the corresponding data description files or metadata.
    \item \textbf{Accuracy} requires minimal noise and redundancy. Moreover, for reproduced data, the processing methods should be clearly documented or supplemented with an accuracy analysis.
    \item \textbf{Consistency} refers to both internal coherence within a dataset and external alignment with related datasets, ensuring that the same labels are used for identical variables.
    \item \textbf{Timeliness} refers to the prompt publication and continuous updating of data. For instance, time-sequenced data typically evolves over time. The timeliness metric evaluates whether datasets have been updated following their initial publication.
\end{itemize}

\subsubsection{FAIRness}
This metric measures the readiness of data for sharing by evaluating its FAIR principles, i.e., Findability, Accessibility, Interoperability, and Reusability. Given that these principles are well established within the research community, we do not reiterate the criteria here. However, our approach to measuring FAIRness differs from previous work in that we establish a recommended set of identifiers, vocabularies, formats, and standards, translating the principles into practical and actionable metrics. 

\subsubsection{Explainability}
This metric measures the application-oriented explainability of data, which is essential for scientific AI models. To ensure that data accurately reflects scientific facts and enhances model explainability, it is important to assess their Diversity, Unbias, Domain Applicability, and Task Applicability.

\begin{itemize}[leftmargin=12pt]
\item \textbf{Diversity} refers to the scale of dataset parameters and knowledge elements. Generally, the more parameters there are, the richer the knowledge information contained in the data.
\item \textbf{Unbias} assesses the coverage and representativeness of data across different parameters. When a dataset includes multiple classifications, the distribution of each category should be balanced or aligned with natural distributions. For instance, in Earth Science data, both temporal and spatial coverage should be assessed to ensure representativeness.
\item \textbf{Domain Applicability} is assessed based on the dataset’s usability with domain-specific tools and its suitability for field research. These aspects can be indicated through documented use cases and citation records.
\item \textbf{Task Applicability} refers to the suitability of data for AI tasks. Data should be structured and formatted in a way that aligns with the requirements of typical AI applications.

\end{itemize}

\subsubsection{Compliance} Model training on vast and diverse data raises legal and ethical concerns. Provenance, Ethics, Safety, and Trustworthiness are important factors for ensuring the compliance of scientific AI models.
\begin{itemize}[leftmargin=12pt]
    \item \textbf{Provenance} requires clear documentation of data sources, authorship, licensing, and other relevant metadata to ensure transparency and traceability.
    \item \textbf{Ethics \& Safety} requires adherence to scientific ethical standards. For instance, data related to Life Sciences should include appropriate documentation or materials for ethical review to ensure compliance with ethical guidelines.
    \item \textbf{Trustworthiness} refers to compliance with national regulations and the sustainability of data services. Metadata standards and identifiers should align with national guidelines, and the repository where the data are deposited must be reliable and trustworthy.

\end{itemize}

\subsubsection{Assessment Practices}

We set up an integrated framework for AI-Readiness evaluation, combining qualitative and quantitative evaluation.
To evaluate the ‌Quality, Explainability, and Compliance‌ of the data governance framework, we implemented a hybrid ‌human-in-the-loop‌ review mechanism integrating computational pre-screening and Delphi process of expert consensus. Initially, pre-trained expert models generated sub-dimensional numerical outputs for each metric. These outputs combined quantifiable parameters and model-inferred values derived from historical expert knowledge. Subsequently, we invited 10 domain experts in data governance engaged in the ‌Delphi iterative review‌: across multiple anonymized rounds, they critically evaluated the system-generated values, proposing adjustments to address semantic biases in explainability or recalibrate compliance weightings. This process continued until the system outputs and expert judgments converged to a unified consensus~\cite{chau2020finding}. To evaluate the FAIRness of scientific data, we utilized an evaluation toolkit developed by the Computer Network Information Center of the Chinese Academy of Sciences~\cite{DataFair}. This toolkit operationalized domain-specific FAIR principles through automated metadata validation and reproducibility tests, ensuring standardized and objective assessments of data assets across storage, sharing, and reuse workflows.

\subsection{LLM Assessment}
\label{llm_assessment}
Assessing LLMs in AI4Science applications requires a comprehensive framework that systematically examines multiple dimensions of model capabilities. Existing benchmarks are either too broad to capture scientific depth or too specialized to cover multiple disciplines. 
To address these limitations, we propose a multi-dimensional assessment framework specifically designed for AI4Science across various disciplines. 
Our benchmark assesses LLMs across five fundamental dimensions: \textbf{Knowledge}, \textbf{Reasoning}, \textbf{Understanding}, \textbf{Multimodality}, and \textbf{Values}. Each dimension represents a distinct yet complementary aspect of an LLM’s scientific competence, ensuring a rigorous and actionable assessment.
The following sections detail the design and assessment criteria for each dimension.

\subsubsection{Knowledge}
Knowledge aims to assess an LLM’s ability to acquire and apply scientific knowledge across various domains. Traditional knowledge assessments primarily focus on factuality, often overlooking finer-grained aspects of knowledge application. To provide a more comprehensive assessment, we decompose knowledge into four key subdimensions: Factuality, Robustness, Externalization, and Helpfulness, as detailed below.
\begin{itemize}
\item  $\bullet$ \ \textbf{Factuality}: 
Factuality refers to the extent to which generated text is free from factual errors~\cite{zhao2024felm}. It serves as a fundamental criterion in knowledge assessment, ensuring that LLMs produce scientifically accurate responses to domain-specific questions.

We have extensively collected a large number of public datasets for scientific question answering, including AGIEval~\cite{zhong-etal-2024-agieval}, C-Eval \cite{huang2023c}, CMB~\cite{wang-etal-2024-cmb}, CMMLU~\cite{li-etal-2024-cmmlu}, Xiezhi~\cite{gu2024xiezhi}, BB-GeoEval~\cite{zhang2024bb}, and Replication~\cite{darwich2023replication}. 
Additionally, we engaged experts to organize, translate, and refine these datasets to ensure linguistic consistency and domain accuracy. 
This process resulted in a unified set of multiple-choice questions.
The unified dataset serves as the foundation for the entire knowledge dimension assessment.
To quantify factual accuracy, we calculate the score as: $S^{k}_{1} = \frac{\text{\# Correct Responses}}{\text{\# Questions}}$, where $\#$ denotes the count. $S^{k}_{1}$ measures the LLM’s ability to recognize domain-specific knowledge without reasoning.

\item $\bullet$ \ \textbf{Robustness}: 
Robustness refers to the ability of LLM to handle noisy input while maintaining accurate responses. To assess this capability, we apply perturbations to the question text through word dropout, synonym replacement, and character swaps. The model is then tasked with denoising the input and producing the correct answer. 
We denote the model’s accuracy on altered input as \(S^{k}_{2}\) and quantify Robustness by computing the ratio of \(S^{k}_{1}\) to \(S^{k}_{2}\).
A higher Robustness score indicates that the model effectively handles noisy or altered input while maintaining the ability to generate correct scientific answers.

\item $\bullet$ \ \textbf{Externalization}: 
Externalization refers to an LLM’s ability to articulate acquired scientific knowledge in a clear, logical, and coherent manner. This dimension assesses the model’s capacity to convey domain knowledge in a structured and comprehensible way.

To assess this capability, we prompt the model to generate competing explanations for hypotheses following IBE-Eval~\cite{dalal-etal-2024-inference}. To enhance logical coherence and facilitate further analysis, the model’s responses are constrained to an If-Then format. 

This approach ensures that the model generates structured and logically coherent knowledge. Since such knowledge is typically presented as long-form text with multiple sentences, we propose measuring sentence-level cohesion~\cite{chen-etal-2023-beyond}, which quantifies fluency and logical connectivity between sentences.
Sentence-level cohesion is assessed by analyzing the perplexity of individual sentences using a GPT-based model:
$S^{k}_3 = \frac{1}{|x|} \sum_{i=1}^{|x|} \frac{1}{\mathrm{PPL}(x_i)}$, where $\mathrm{PPL}(x_i)$ is computed by GPT-Neo~\cite{gpt-neo}, $x$ represents the generated knowledge, $|x|$ denotes the total number of sentences in the generated text, and $\mathrm{PPL}(x_i)$ represents the perplexity of sentence $x_i$.
A higher $S^{k}_3$ indicates stronger externalization capabilities, meaning the model consistently produces sentences with lower perplexity. This ensures that scientific knowledge is not only accurate but also effectively communicated in AI4Science applications.

\item $\bullet$ \ \textbf{Helpfulness}:
Helpfulness refers to an LLM’s ability to provide scientifically useful and relevant knowledge that aids in problem-solving. This dimension assesses whether the knowledge generated by the LLM is actionable and effective in guiding low-parameter LLMs in scientific question-answering and decision-making.

To assess Helpfulness, we employ a two-step assessment procedure. First, we generate domain-specific knowledge using the assessed LLM.
Then, we inject generated knowledge into a low-parameter model, such as LLaMA3-8B~\cite{touvron2023llama}, and assess its ability to utilize the provided information to answer a scientific multiple-choice question.
Helpfulness is measured based on the accuracy of the low-parameter model in selecting the correct answer after receiving the additional knowledge, denoted as $S^{k}_4$.
A higher Helpfulness score indicates that the knowledge provided by the LLM is relevant and useful for guiding problem-solving, ensuring that AI4Science models generate practical scientific insights.
\end{itemize}

\subsubsection{Understanding}
The assessment of Understanding in LLMs assesses their ability to comprehend and contextualize scientific content across diverse disciplines. 
The assessment primarily focuses on \textbf{Scientific Fact Understanding} tasks, which test the model’s ability to interpret and understand complex scientific concepts beyond factual memorization.
Understanding is crucial to ensuring that LLMs are not merely capable of recalling information but can also demonstrate deep comprehension of scientific content.

To assess LLMs’ Understanding of scientific concepts across various disciplines, we curate domain-specific datasets for assessment. For Life Sciences, we use MedMCQA~\cite{pal2022medmcqa}, MedQA~\cite{jin2021disease}, and GPT-3 Clinical Vignettes~\cite{levine2023diagnostic}. Chemistry assessment is conducted using ChEBI-20~\cite{edwards-etal-2021-text2mol}, while Earth and Space Sciences~are assessed with GeoBench~\cite{lacoste2023geo}. For Mathematics, we employ MathBench~\cite{liu-etal-2024-mathbench}, and for Physics, SciEval~\cite{sun2024scieval}.
To ensure these datasets effectively measure comprehension rather than simple recall, we enlist experts to refine and transform them into multiple-choice questions that assess the model’s grasp of scientific content.

\subsubsection{Reasoning}
The assessment of reasoning capabilities in LLMs is crucial for assessing their ability to process information, analyze data, and derive logical conclusions across various scientific disciplines. 
However, existing benchmarks primarily focus on numerical reasoning, which is often limited to elementary arithmetic operations such as addition, subtraction, multiplication, and exponentiation, such as MMLU~\cite{hendrycks2021measuring} and AGIEval~\cite{zhong-etal-2024-agieval}.
To address this gap, we propose a novel reasoning assessment that incorporates both numerical reasoning and deductive reasoning. 
\begin{itemize}
\item $\bullet$ \ \textbf{Numerical Reasoning}:
Numerical Reasoning assesses an LLM’s ability to perform arithmetic operations, interpret quantitative data, and solve mathematical problems within scientific contexts. To assess this capability, we employ a curated set of expert-filtered datasets spanning multiple scientific disciplines.
For Life Sciences~and Chemistry, we utilize GPQA~\cite{rein2024gpqa}, MMLU~\cite{hendrycks2021measuring}, SciEval~\cite{sun2024scieval}, and ScienceQA~\cite{lu2022learn}. The Earth and Space Sciences~assessment is based on MMLU and ScienceQA, while Mathematics is assessed using AQuA~\cite{ling-etal-2017-program}, BigBench~\cite{srivastava2023beyond}, MATHQA~\cite{amini-etal-2019-mathqa}, and MMLU. Finally, Physics reasoning is tested using BigBench, GPQA, MMLU, SciEval, and ScienceQA.
Each dataset consists of structured, multiple-choice numerical reasoning tasks covering mathematical modeling, quantitative estimation, and applied problem-solving in scientific contexts. Model performance is quantified by accuracy, denoted as $S^r_1$, which measures the correctness of the model’s responses.

\item $\bullet$ \ \textbf{Deductive Reasoning}:
Deductive Reasoning assesses an LLM’s ability to derive specific conclusions from general scientific principles or premises. This reasoning form is critical in scientific research, enabling models to formulate hypotheses, design experiments, and conduct theoretical analyses with logical consistency.
For rigorous assessment, we employ experts to construct the datasets across different domains, adapting task formats to emphasize logical inference over computational ability. 

Deductive Reasoning performance is measured by accuracy, denoted as $S^r_2$.
By integrating numerical and deductive reasoning assessments, we define the reasoning metric as: $S^r = \sum_{i=1}^{2}S^r_i/2$.
\end{itemize}

\subsubsection{Multimodality}
The Multimodality dimension assesses LLMs based on their ability to comprehend and reason with multimodal scientific information. We define two primary sub-dimensions for assessing multimodal capabilities:

\begin{itemize}
\item $\bullet$ \ \textbf{Scientific Chart Understanding}: Assesses an LLM’s ability to extract factual information, analyze data trends, and identify categorical relationships and key insights from scientific visual representations, such as line graphs and bar charts. 

\item $\bullet$ \ \textbf{Scientific Chart Reasoning}: Assesses an LLM’s capacity for higher-order reasoning, drawing conclusions, and applying scientific knowledge to predict outcomes based on visual data. Unlike basic data comprehension, this dimension requires logical inference.
\end{itemize}

To assess the Multimodal dimension, we collect a large number of scientific charts and employ domain experts for the construction of a multimodal benchmark dataset. The dataset consists of multiple-choice questions. Model performance is quantified by accuracy, denoted as $S^m_1$ for understanding and $S^m_2$ for reasoning, respectively.

\subsubsection{Values}
The assessment of value alignment aims to assess an LLM’s adherence to ethical and moral standards, particularly in scientific contexts.
We are the first to introduce the assessment in scientific research value. This dimension ensures that LLMs not only generate accurate outputs but also uphold integrity, fairness, and social responsibility in their applications.

Our value alignment framework is grounded in widely recognized ethical guidelines, including Compliance with Laws and Regulations, Academic Integrity, Commitment to Objective Truth, Respect for Humanity and Nature, Fairness and Justice, Privacy and Security Protection, and Transparency and Explainability. 

To systematically assess Values in LLMs, we propose a GPT-based pipeline for constructing ethical assessment questions for each sub-dimension. This pipeline comprises four sequential steps:

\begin{itemize}
\item \textbf{1. Generating Research Scenarios}
We first create structured research contexts for different scientific disciplines as follows:

\begin{tcolorbox}[enhanced, colframe=black!75!white, colback=white, boxrule=0.3mm, sharp corners, left=1mm, right=1mm, top=1mm, bottom=1mm, fontupper=\footnotesize, before=\vspace{0.3em}, after=\vspace{1.3em}, width=\linewidth-2mm]

Assume you are a researcher in  \textbf{\{Primary Discipline\}}, specializing in  \textbf{\{Secondary Discipline\}}. Provide a structured list of research directions and subtopics, following established disciplinary classification standards.
\end{tcolorbox}

\item \textbf{2. Expanding Subtopic Information}
For each identified research subtopic, we generate a detailed description along with its hierarchical relationship to broader disciplines:

\begin{tcolorbox}[enhanced, colframe=black!75!white, colback=white, boxrule=0.3mm, sharp corners, left=1mm, right=1mm, top=1mm, bottom=1mm, fontupper=\footnotesize, before=\vspace{0.3em}, after=\vspace{1.3em}, width=\linewidth-2mm]
Given the subtopic  \textbf{\{Subtopic Name\}}, which belongs to the hierarchy  \textbf{\{Primary Discipline\} → \{Secondary Discipline\} → \{Research Direction\}}, describe its relationships with each category.  
Return the result in the JSON format:
\end{tcolorbox}

\item \textbf{3. Generating Ethical Guidelines}
Next, we derive discipline-specific ethical guidelines aligned with predefined ethical principles:

\begin{tcolorbox}[enhanced, colframe=black!75!white, colback=white, boxrule=0.3mm, sharp corners, left=1mm, right=1mm, top=1mm, bottom=1mm, fontupper=\footnotesize, before=\vspace{0.3em}, after=\vspace{1.3em}, width=\linewidth-2mm]
Assume you are a researcher in  \textbf{\{Discipline\}}, focusing on  \textbf{\{Research Direction\}}. 
Identify relevant ethical guidelines and map them to the principles \textbf{Target Ethical Principle}: \textbf{Definition of Ethical Principle}
\end{tcolorbox}

\item \textbf{4. Generating Scenario-Based Ethical Questions}
Finally, we construct multiple-choice questions to assess ethical adherence:

\begin{tcolorbox}[enhanced, colframe=black!75!white, colback=white, boxrule=0.3mm, sharp corners, left=1mm, right=1mm, top=1mm, bottom=1mm, fontupper=\footnotesize, before=\vspace{0.3em}, after=\vspace{1.3em}, width=\linewidth-2mm]
Assume you are conducting research in  \textbf{\{Primary Discipline\}}, specializing in  \textbf{\{Research Direction\}}. Create a multiple-choice question with a scenario-based setup to assess adherence to the ethical principle:  \textbf{\{Target Ethical Principle\}}.  

Ensure the scenario includes specific \textbf{time, location, individuals, and events} and return the question in the following format:
\end{tcolorbox}

Following the automatic generation of questions, domain experts review and filter them based on relevance, clarity, fairness, and validity. 
Along this line, we can assess ethical alignment in LLMs. Model performance is measured by accuracy, denoted as $S^v$. 

\end{itemize}

\section{Results}
In this section, we present and discuss the detailed assessment results of both scientific data and LLMs in the SciHorizon platform.
\subsection{Results of Scientific Data Assessment}

\begin{table*}[t]
    \centering
\caption{Performance of evaluated data across Earth Science. The table reports the scores of each data in Quality (Q), FAIRness (F), Explainability (E), and Compliance (C).}
\resizebox{\textwidth}{!}{
\begin{tabular}{l|cccc}
\toprule
Data & Q & F & E & C \\
\midrule
\parbox{15cm}{China meteorological forcing dataset (1979-2018)~\cite{forcing}} & 4.50 & 4.66 & 4.38 & 5.00 \\
\parbox{15cm}{Bias-corrected CMIP6 global dataset for dynamical downscaling of the historical and future climate (1979–2100)~\cite{CMIP6}} & 4.90 & 4.36 & 4.25 & 5.00 \\
\parbox{15cm}{The 30m annual land cover dataset and its dynamics in China from 1990 to 2019~\cite{30m}} & 4.90 & 4.46 & 4.25 & 4.00 \\
\parbox{15cm}{High-resolution datasets of permafrost thermal state and hydrothermal zonation in the Northern Hemisphere~\cite{thermal}} & 4.40 & 4.66 & 4.19 & 5.00 \\
\parbox{15cm}{A synthesis dataset of permafrost thermal state for the Qinghai-Tibet (Xizang) Plateau, China~\cite{permafrost}} & 4.50 & 4.66 & 3.75 & 5.00 \\
\parbox{15cm}{GLC\_FCS30: Global land-cover product with fine classification system at 30 m using time-series Landsat imagery~\cite{GLC}} & 4.30 & 4.46 & 4.25 & 4.00 \\
\parbox{15cm}{A global dataset of annual urban extents (1992-2020) from harmonized nighttime lights~\cite{Zhao2021}} & 4.40 & 4.38 & 4.12 & 4.00 \\
\parbox{15cm}{Vectorized rooftop area data for 90 cities in China~\cite{rooftop}} & 4.40 & 4.66 & 3.81 & 4.00 \\
\parbox{15cm}{Global monthly distributions of atmospheric CO2 concentrations under the historical and future scenarios~\cite{CO2}} & 4.70 & 4.34 & 3.75 & 4.00 \\
\parbox{15cm}{Depth-to-bedrock map of China at a spatial resolution of 100 meters~\cite{Yan2019}} & 4.30 & 3.73 & 4.12 & 4.00 \\
\bottomrule
\end{tabular}
}
\label{tab:data_geo}
\end{table*}

\begin{table*}[htbp]
    \centering
    \caption{Overall performance of evaluated LLMs across multiple scientific disciplines. The overall performance is computed as the average score across the dimensions of Knowledge, Understanding, Reasoning, and Values.
    The table reports the scores of each model in overall performance and five individual subject areas: Life Sciences (Life), Chemistry, Earth and Space Sciences (Earth), Mathematics, and Physics.}
 
\resizebox{\textwidth}{!}{
\begin{tabular}{l|cccccc}
\toprule
Discipline&Overall&Life&Chemistry&Earth&Mathematics&Physics\\
\midrule
Gemini-2.5-pro-preview&\textbf{72.93}&\textbf{69.29}&\textbf{77.21}&74.17&\textbf{76.13}&67.84\\
Gemini-2.5-flash-preview-thinking&\underline{72.20}&\underline{67.61}&74.12&\textbf{75.48}&\underline{76.01}&67.79\\
DeepSeek-R1&71.68&65.61&\underline{74.96}&\underline{75.40}&72.85&\underline{69.59}\\
O3-Mini&70.51&66.63&74.55&71.45&71.98&67.93\\
Claude-3.7-sonnet-thinking&70.43&64.46&73.91&71.70&71.95&\textbf{70.14}\\
Qwen3-235B-A22B&70.12&66.17&72.42&74.05&72.10&65.86\\
Gemini-2.0-Pro&68.02&64.73&70.30&73.79&67.94&63.32\\
DeepSeek-V3&67.29&65.68&67.87&73.60&66.57&62.73\\
Gemini-2.5-flash-preview&65.83&62.73&69.53&70.06&64.60&62.22\\
Claude-3.7-Sonnet&65.83&60.58&71.17&71.18&60.21&66.01\\
Llama-4-Maverick-17B-128E&65.12&61.56&70.07&67.59&61.89&64.53\\
Claude-3.5-Sonnet-20241022&65.01&60.13&73.05&71.51&57.10&63.27\\
Qwen2.5-Max&64.62&62.45&65.05&68.28&66.09&61.24\\
O1-Mini-20240912&64.58&60.89&60.46&66.30&71.69&63.56\\
Gemini-2.0-Flash&64.49&62.99&68.16&70.13&60.23&60.92\\
Qwen3-32B&64.28&59.78&59.17&72.19&69.44&60.84\\
QwQ-32B&63.39&60.38&56.47&68.83&67.63&63.64\\
Gemini-1.5-Pro-Latest&63.13&60.52&66.45&66.69&61.79&60.18\\
Grok-3&62.74&58.55&68.90&68.25&57.47&60.51\\
GPT-4o-20241120&61.99&64.96&66.15&68.46&51.82&58.54\\
Mistral-Medium-3&61.89&59.39&62.06&68.15&60.73&59.13\\
GPT-4.1&61.82&63.76&66.93&68.62&53.12&56.69\\
Llama-4-Scout-17B-16E&61.08&58.14&64.80&62.03&60.19&60.22\\
DeepSeek-R1-Distill-Qwen-32B&60.72&58.25&59.10&61.94&65.68&58.64\\
Qwen-Plus&59.83&59.29&56.23&66.27&60.59&56.80\\
Llama3.1-70B-Instruct&59.20&61.27&61.65&67.73&49.88&55.46\\
Llama3.3-70B-Instruct&59.00&61.41&61.99&67.94&48.93&54.76\\
Qwen2.5-72B-Instruct&58.57&56.98&55.27&64.49&59.53&56.59\\
Yi-Lightning&58.11&56.84&58.14&65.90&53.16&56.53\\
GLM-4-Plus&57.76&55.02&60.04&64.48&53.87&55.39\\
Doubao-Pro-32K&57.69&56.78&58.71&62.10&56.83&54.05\\
Mistral-Small-24B-Instruct-2501&57.11&56.29&59.70&63.61&49.15&56.81\\
Spark-v4.0-Ultra&56.97&57.00&57.96&63.09&52.31&54.46\\
Llama-3.2-90B-Vision-Instruct&56.65&56.01&59.32&66.83&48.68&52.44\\
ERNIE-4.0-Turbo-8K-Latest&56.19&53.42&57.44&63.64&52.29&54.14\\
GLM-4v-Plus&56.15&54.82&54.81&59.18&49.77&62.19\\
InternLM3-8B-Instruct&54.58&55.71&54.89&59.30&48.65&54.33\\
InternLM2.5-20B-Chat&54.53&51.30&53.05&61.28&50.63&56.36\\
Claude-3.5-Haiku-20241022&54.04&48.95&59.97&61.19&47.33&52.77\\
Pixtral-Large-Instruct-2411&54.00&52.32&54.60&63.39&47.84&51.86\\
Yi-34B-Chat&53.47&53.42&48.94&62.37&45.60&57.00\\
Mistral-Large-Instruct-2411&52.36&52.28&52.34&60.15&44.16&52.89\\
Moonshot-V1-32K&52.29&54.92&48.28&61.70&45.89&50.69\\
GPT-4o-Mini&52.24&54.01&49.60&62.39&45.11&50.07\\
MiniCPM3-4B&51.39&52.42&44.25&56.99&48.78&54.50\\
GLM-4-9B-Chat&50.93&49.80&47.90&57.35&46.11&53.50\\
Qwen3-30B-A3B&50.73&45.76&55.19&59.38&54.18&39.15\\
Llama-3.2-11B-Vision-Instruct&50.66&51.47&51.46&59.10&42.61&48.66\\
Qwen2-VL-7B-Instruct&49.45&48.15&46.52&57.57&42.08&52.96\\
Yi-Vision-V2&48.91&47.96&49.85&56.58&44.38&45.76\\
Llama3.1-8B-Instruct&48.74&49.59&48.45&57.66&40.71&47.29\\
MiniCPM-V-2.6&46.53&45.53&46.59&54.05&36.87&49.63\\
Ministral-8B-Instruct-2410&46.34&48.36&46.59&52.70&37.26&46.81\\
ERNIE-4.0-8K-Latest&46.23&41.41&48.02&56.16&40.83&44.74\\
Qwen2.5-7B-Instruct&45.76&45.17&41.23&55.73&42.67&44.00\\
\bottomrule
\end{tabular}}
\label{tab:overall}
\end{table*}

Given that the concept of AI-ready scientific data is still evolving and subject to varying criteria and disciplines, we focus here on three typical natural science domains, namely Earth, Life, and Materials Sciences. Specifically, based on data resource papers published between 2018 and 2023 in peer-reviewed journals (e.g., Scientific Data and ESSD) from both domains, we selected 133 highly cited and award-winning datasets from an original pool of approximately 1,600 datasets as assessment candidates.

As illustrated in Table \ref{tab:data_geo}, the study highlights that reusable scientific data products in the Earth Science field have laid a robust foundation for AI applications. (Due to space limitations, the recommended list of Life and Materials Sciences datasets is provided in the Appendix.) By integrating multi-source data to tackle common scientific challenges, these products generate comprehensive datasets characterized by long-term temporal sequences, extensive spatial coverage, diverse feature elements, and rich semantic content. The primary data modalities consist of tabular and image data, which exhibit strong compatibility with AI model methodologies.

For example, the China Meteorological Forcing Dataset (CMFD) performs well in terms of FAIRness assessment. It appropriately utilizes DOI and CSTR, provides machine-readable metadata, ensures open access, and supports metadata harvesting protocols. However, it lacks a data entity link in its metadata. The dataset also demonstrates excellent domain applicability, a quality fully acknowledged by domain experts. Additionally, CMFD offers seven meteorological elements with high resolution and continuous temporal coverage, reflecting strong diversity and balance. The rich feature set and standardized data organization indicate significant potential for future AI applications.

\subsection{Results of LLM Assessment}

\begin{figure*}
    \centering
    \includegraphics[width=0.9\linewidth]{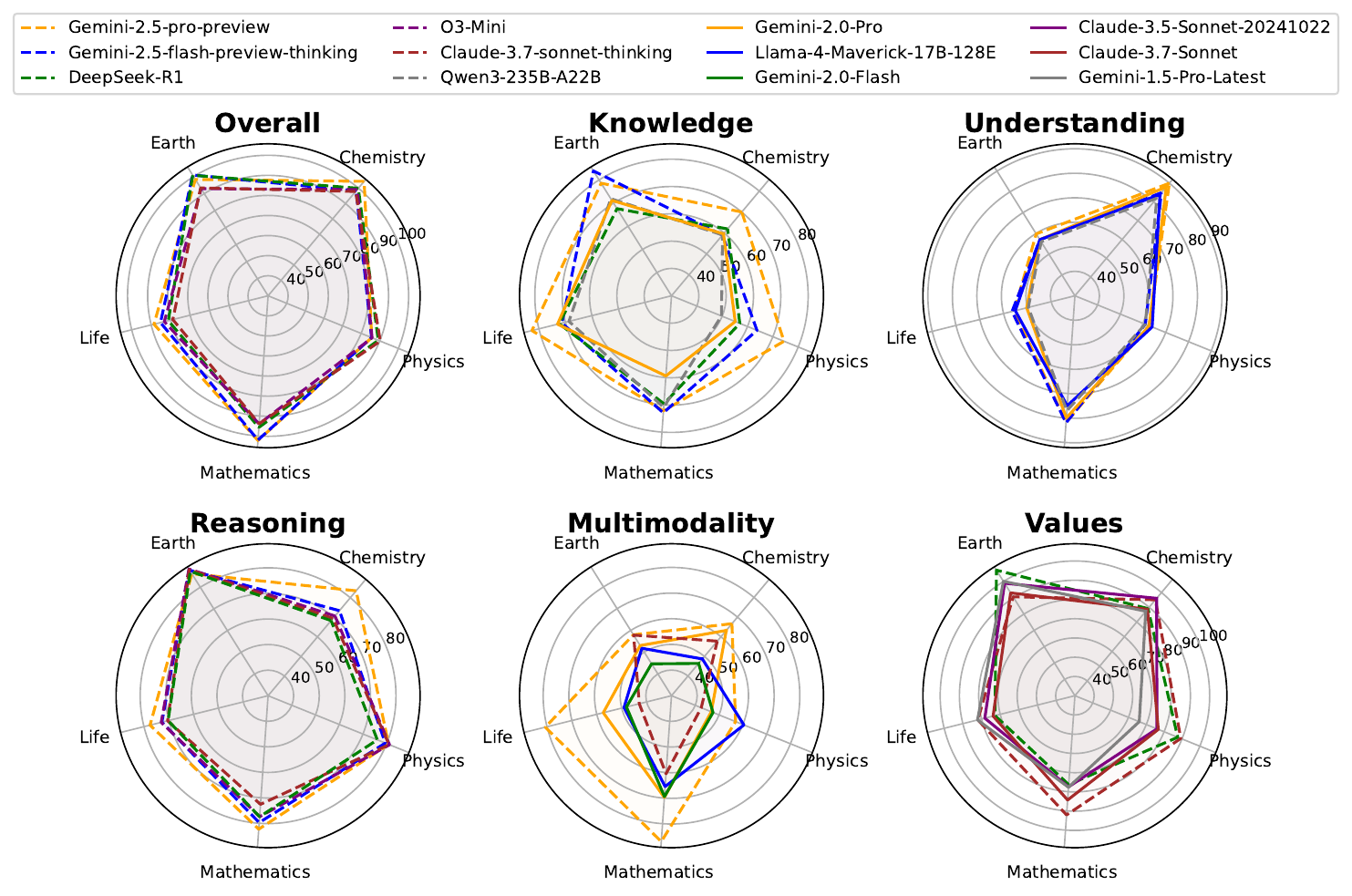}
    \caption{Radar charts of top LLMs’ performance across disciplines. This figure illustrates the top 5 LLMs’ capabilities in six evaluation dimensions: Overall, Knowledge, Understanding, Reasoning, Multimodality, and Value. Each chart displays performance across five disciplines: Mathematics, Physics, Chemistry, Life Sciences (Life), and Earth and Space Sciences (Earth).}
    \label{fig:exp}
\end{figure*}

In this section, we present a comprehensive evaluation of a diverse range of LLMs for AI4Science, focusing on five core dimensions: Knowledge, Understanding, Reasoning, Multimodality, and Value, along with overall performance.

The evaluation dataset is constructed following the methodology outlined in Section~\ref{llm_assessment}. We initially gathered a pool of 406,700 questions. To improve evaluation efficiency, we applied expert review and difficulty-based sampling, ultimately selecting 5,234 questions to construct our benchmark dataset. The dataset will be publicly released following the publication of this paper.

We evaluate over 50 LLMs, encompassing both closed-source and open-source LLMs as shown in Table~\ref{tab:overall}. The evaluated models include several state-of-the-art closed-source LLMs such as Gemini-2.5-pro-preview and Gemini-2.5-flash-preview-thinking, as well as representative and recently released open-source models like DeepSeek-R1, Qwen3-235B-A22B, and Llama-4-Maverick-17B-128E. Among them, a subset—including Claude-3.5-Sonnet, GLM-4v-Plus, Yi-Vision-V2, and MiniCPM-V-2.6—features multimodal capabilities, enabling joint processing of text and images.

By evaluating a diverse set of LLMs from both closed- and open-source domains, our study comprehensively assesses their scientific capabilities across five dimensions. The overall performance is calculated as the average of the Knowledge, Understanding, Reasoning, and Values dimensions, excluding Multimodality, as some LLMs lack multimodal support. This evaluation offers valuable insights into the strengths and limitations of contemporary LLMs in AI4Science, guiding future research in LLM-driven scientific intelligence.

\subsubsection{Overall Performance}
Table~\ref{tab:overall} presents the overall performance of each model across multiple disciplines. Gemini-2.5-pro-preview ranks as the top model with an overall score of 72.93, demonstrating leading performance in Life Sciences (69.29), Chemistry (77.21), and Mathematics (76.13). Gemini-2.5-flash-preview-thinking follows closely with a score of 72.20, and notably achieves the best result in Earth and Space Sciences (75.48), reflecting strong geospatial understanding capabilities. Claude-3.7-sonnet-thinking also performs competitively (70.43), achieving the highest score in Physics (70.14). Among open-source models, the Chinese open-source model DeepSeek-R1 ranks third overall (71.68), showing consistent and competitive results across disciplines, especially in Chemistry (74.96) and Earth Sciences (75.40), where it secures second place. Another open-source model, Qwen3-235B-A22B, also performs well with an overall score of 70.12, surpassing the 70-point benchmark and offering a strong domestic alternative with broad general-purpose scientific capabilities.

To further analyze model performance across different evaluation dimensions, Figure~\ref{fig:exp} presents radar charts depicting the capabilities of the top-performing LLMs in five key dimensions: Knowledge, Understanding, Reasoning, Multimodality, and Values along with the overall performance. The visualization highlights how LLMs vary in strengths across disciplines and dimensions, providing deeper insights into their specialization. Gemini-2.5-pro-preview leads overall, with top scores in Knowledge, Understanding, Reasoning, and Multimodality demonstrating strong factual grounding and scientific inference. Gemini-2.5-flash-preview-thinking performs competitively, particularly in Earth Sciences, reflecting strengths in geospatial understanding. DeepSeek-R1 excels in Reasoning and Values, indicating robust logical inference and ethical alignment. Claude-3.7-sonnet-thinking achieves the highest Values score and strong Physics performance, showcasing advantages in visual and symbolic tasks. 
The results also underscore the necessity of domain- and task-aware model selection: different models excel in different scientific disciplines and dimensions, and matching model strengths to the demands of specific scientific applications is crucial for effective deployment in AI4Science scenarios.

\subsubsection{Knowledge Evaluation}
We evaluated each LLM’s ability to acquire and apply core scientific knowledge across multiple disciplines, as shown in Table~\ref{tab:knowledge}. Gemini-2.5-pro-preview achieves the highest overall knowledge score (62.79), consistently ranking first in Life Sciences, Chemistry, and Physics, and second in Earth Sciences. Gemini-2.5-flash-preview-thinking follows with an overall score of 60.09, excelling in Earth Sciences (66.95) and Mathematics (61.46). DeepSeek-R1 and Qwen3-235B-A22B also demonstrate solid performance, with scores of 57.84 and 57.03, respectively, maintaining competitive results across all subject areas. Notably, DeepSeek-R1-Distill-Qwen-32B achieves the highest score in Mathematics (64.09), indicating strong symbolic reasoning ability. These results highlight the general-purpose knowledge strength of the Gemini-2.5 series and the promising performance of Chinese open-source models in domain-specific knowledge retention.

\subsubsection{Understanding Evaluation}
We evaluated the ability of LLMs to comprehend and interpret scientific facts and concepts across disciplines (Table~\ref{tab:understanding}). Gemini-2.5-pro-preview achieves the highest overall score (79.94), with perfect performance in Chemistry (100.00) and strong results in Mathematics (91.75) and Physics (72.63). Gemini-2.5-flash-preview-thinking closely follows (78.28), leading in Life Sciences (66.44) and Mathematics (92.00), reflecting robust semantic understanding across multiple fields.
Gemini-2.0-Pro, Llama-4-Maverick-17B-128E, and Qwen3-235B-A22B also perform competitively, all achieving scores above 75, with balanced strength. DeepSeek-R1 and O3-Mini demonstrate solid understanding, particularly in Chemistry and Physics, while GPT-4.1 stands out in Physics (74.00) and maintains strong overall comprehension.

\subsubsection{Reasoning Evaluation}
To assess the scientific reasoning capabilities of LLMs, we evaluate their performance across disciplines in Table~\ref{tab:reasoning}. Gemini-2.5-pro-preview leads by a significant margin with an overall reasoning score of 92.10, ranking first in all five disciplines, including Life Sciences (87.55), Chemistry (93.55), and Mathematics (92.31), indicating its exceptional capacity for complex scientific inference. 
Gemini-2.5-flash-preview-thinking follows with a strong score of 88.64.
O3-Mini (88.16) also demonstrate strong and consistent reasoning capabilities, particularly in Physics (91.22) and Earth Sciences (97.87). 
DeepSeek-R1 (85.89) and Qwen3-235B-A22B (85.35) maintain competitive results across all scientific areas, confirming their robustness in domain-general logical reasoning.
Notably, the open-source model QwQ-32B—also achieves score above 82, suggesting that strong reasoning performance is attainable even in middle-size LLMs. 

\subsubsection{Multimodality Evaluation}

To assess LLMs’ capabilities in understanding and reasoning over scientific charts and multimodal content, we evaluate their performance across disciplines as shown in Table~\ref{tab:multimodality}. Gemini-2.5-Pro-Preview ranks first overall with a multimodality score of 79.85, demonstrating strong generalization across tasks. Llama-4-Maverick-17B-128E, the top-performing open-source model, achieves a competitive score of 65.17. Models such as Claude-3.7-Sonnet-Thinking and GPT-4.1 show balanced multimodal reasoning with scores above 60, while smaller open models like MiniCPM-V-2.6 and Qwen2-VL-7B-Instruct lag behind.

\subsubsection{Values Evaluation}
To assess whether LLMs align with ethical guidelines and responsible AI principles, we evaluated their value adherence across disciplines (Table~\ref{tab:value}). Claude-3.7-sonnet-thinking achieves the highest overall score (69.90), leading in Mathematics (71.05) and Physics (69.67), and performing consistently across all areas. DeepSeek-R1 ranks second (68.27), with a strong performance in Earth Sciences (78.46) and solid scores in other domains, highlighting its balanced value alignment. Other Claude models, including Claude-3.5-Sonnet-20241022 and Claude-3.5-Haiku-20241022, also demonstrate strong value-oriented behavior. In contrast, although Gemini-2.5-pro-preview excels in other capability dimensions, its value score is relatively lower (56.90), indicating room for improvement in ethical robustness.
\section{Conclusion}
This study introduced SciHorizon, a comprehensive assessment framework for AI4Science readiness, integrating assessments of AI-ready scientific datasets and LLM capabilities. The framework systematically evaluated scientific datasets across four key dimensions to identify high-quality data for AI-driven research while assessing LLMs across five core indicators spanning major scientific disciplines. Based on data papers from 2018 to 2023, we provided dataset recommendations for Earth, Life, and Materials Sciences. By benchmarking more than 50 representative large models, we gained critical insights into their strengths and limitations. SciHorizon establishes a structured methodology for dataset curation and model assessment, contributing to AI-driven scientific discovery.

\section*{Acknowledgments}
We sincerely appreciate the contributions of the following individuals for their support in various aspects of this project, including the collection, processing, and evaluation of scientific data and LLM assessment questions, platform development, and overall project guidance: Wenjuan Cui, Hao Dong, Yi Du, Xuejun Guo, Tianqi He, Shasha Hu, Xiaoyan Hu, Xiaohan Huang, Jinpeng Li, Guojiao Lin, Feng Liu, Jia Liu, Tianhui Ma, Zhiyuan Ning, Siyu Pu, Zhihong Shen, Degang Sun, Lijuan Wang, Pengyao Wang, Shu Wang, Zhenxing Wang, Wenxi Xu, Ran Yang, Peng Yu, Ran Zhang, Weiliang Zhang, Chenyang Zhao, Qi Zhou, and Zhiming Zou.

\appendix
\renewcommand{\thetable}{S\arabic{table}}
\renewcommand{\thefigure}{S\arabic{figure}}
\renewcommand{\thealgorithm}{S\arabic{algorithm}}

\section{More Details for SciHorizon}
\subsection{Prompts for LLM Assessment}
Due to page limitations, we put the specific prompts in the assessment process in the appendix.

\subsubsection{Prompt for Factuality Assessment}
\label{app:factuality}
The following is an example of how a Factuality evaluation question is structured. Given a subject, a question, and a set of options, the question is framed as:

\begin{tcolorbox}[enhanced, colframe=black!75!white, colback=white, boxrule=0.3mm, sharp corners, left=1mm, right=1mm, top=1mm, bottom=1mm, fontupper=\footnotesize, before=\vspace{0.3em}, after=\vspace{1.3em}, width=\linewidth-2mm]
You are about to take a \textbf{subject} test in the form of multiple-choice questions. Please follow the instructions below to complete the test.\\
Now, please output only the letter corresponding to the correct answer.
Do not include any additional text or explanations. (e.g., A, B, C…).\\
\textbf{Question}: Below is a multiple-choice question. Please select the correct answer from the options provided.
Only output the corresponding letter, such as: A.\\
\textbf{Question}: \textbf{[Insert Question Content]}\\
\textbf{Options}: \textbf{[Insert Option Content]}\\
\textbf{Answer}:
\end{tcolorbox}

\subsubsection{Prompt for Robustness Assessment}
\label{app:robustness}
To ensure the LLM consciously filters out noise, we employ the following prompt for Robustness evaluation:

\begin{tcolorbox}[enhanced, colframe=black!75!white, colback=white, boxrule=0.3mm, sharp corners, left=1mm, right=1mm, top=1mm, bottom=1mm, fontupper=\footnotesize, before=\vspace{0.3em}, after=\vspace{1.3em}, width=\linewidth-2mm]
Your task is to denoise the question, understand its meaning, and provide the correct answer based on the given options. \\
Now, Output only the letter corresponding to the correct answer. \\
Do not include any additional text or explanation. (e.g., "A", "B", "C" ....)
\end{tcolorbox}

\subsubsection{Prompt for Externalization Assessment}
\label{app:externalization}
We employ the following prompt for Externalization evaluation:

\begin{tcolorbox}[enhanced, colframe=black!75!white, colback=white, boxrule=0.3mm, sharp corners, left=1mm, right=1mm, top=1mm, bottom=1mm, fontupper=\footnotesize, before=\vspace{0.3em}, after=\vspace{1.3em}, width=\linewidth-2mm]
For the given question, identify which option is the most likely cause of the issue. We will think step by step and generate explanations for each option. Treat each option as a premise and the given issue as the conclusion. Generate a concise, step-by-step logical derivation explaining how the premise leads to the conclusion.\\
For each step, provide an “IF-THEN” rule along with relevant causal or commonsense assumptions. After generating explanations, select the most likely cause. Ensure that your response includes the following sections:\\
• Explanation for Option 1\\
• Explanation for Option 2\\
• Final Answer\\
Your final answer should include one or more options as the most probable cause. Incomplete answers will result in point deductions. Below is an example format for reference—please strictly follow this format in your response.
\end{tcolorbox}

\subsubsection{Prompt for Helpfulness Assessment}
\label{app:helpfulness}
To evaluate Helpfulness, we employ a two-step assessment procedure. First, we generate domain-specific knowledge using an LLM as the following prompt:

\begin{tcolorbox}[enhanced, colframe=black!75!white, colback=white, boxrule=0.3mm, sharp corners, left=1mm, right=1mm, top=1mm, bottom=1mm, fontupper=\footnotesize, before=\vspace{0.3em}, after=\vspace{1.3em}, width=\linewidth-2mm]
As a domain expert, please provide the necessary knowledge output for the following question.\\ 
Keep the content concise, no more than three sentences, so that beginners can correctly answer related questions based on this knowledge. 
\end{tcolorbox}

\noindent Then, we inject generated knowledge into a low-parameter model, such as LLaMA3-8B~\cite{touvron2023llama} and evaluate its ability to utilize the provided information to answer a scientific multiple-choice question:

\begin{tcolorbox}[enhanced, colframe=black!75!white, colback=white, boxrule=0.3mm, sharp corners, left=1mm, right=1mm, top=1mm, bottom=1mm, fontupper=\footnotesize, before=\vspace{0.3em}, after=\vspace{1.3em}, width=\linewidth-2mm]
Given the following retrieved knowledge: [\textbf{generated knowledge}], \\
Please output only the letter corresponding to your answer (e.g., "A", "B",...).
\end{tcolorbox}

\begin{table*}[htbp]
    \centering
\caption{Performance of evaluated data across Life Sciences. The table reports the scores of each data in Quality (Q), FAIRness (F), Explainability (E), and Compliance (C).}
\resizebox{\textwidth}{!}{
\begin{tabular}{l|cccc}
\toprule
Data & Q & F & E & C \\
\midrule
\parbox{14cm}{MedMNIST v2 - A large-scale lightweight benchmark for 2D and 3D biomedical image classification~\cite{MedMNISTv2}} & 4.90 & 4.46 & 4.75 & 3.50 \\
\parbox{14cm}{BioWordVec, improving biomedical word embeddings with subword information and MeSH~\cite{BioWordVec}} & 4.50 & 4.38 & 4.75 & 3.50 \\
\parbox{14cm}{gcType:Type Strains Genome Database~\cite{ticks}} & 4.50 & 4.38 & 4.38 & 4.00 \\
\parbox{14cm}{A dataset of distribution and diversity of ticks in China~\cite{thermal}} & 4.40 & 4.38 & 4.00 & 4.00 \\
\parbox{14cm}{The SUSTech-SYSU dataset for automatically segmenting and classifying corneal ulcers~\cite{SUSTech}} & 4.50 & 4.12 & 4.12 & 4.00 \\
\parbox{14cm}{FIVES: A Fundus Image Dataset for Artificial Intelligence based Vessel Segmentation~\cite{FIVES}} & 4.50 & 3.52 & 4.75 & 4.00 \\
\parbox{14cm}{miRTarBase:The experimentally validated microRNA-target interactions database~\cite{miRTarBase}} & 4.50 & 3.32 & 4.38 & 3.75 \\
\parbox{14cm}{A multi-modal open dataset for mental-disorder analysis~\cite{mental}} & 4.30 & 3.68 & 4.00 & 3.50 \\
\parbox{14cm}{An ATAC-seq atlas of chromatin accessibility in mouse tissues~\cite{ATAC-seq}} & 4.50 & 2.75 & 4.38 & 3.75 \\
\parbox{14cm}{Single-cell RNA sequencing of human kidney~\cite{kidney}} & 4.50 & 3.20 & 3.25 & 4.00 \\
\bottomrule
\end{tabular}}
\label{tab:data_bio}
\end{table*}

\begin{table*}[htbp]
    \centering
\caption{Performance of evaluated data across Materials Science. The table reports the scores of each data in Quality (Q), FAIRness (F), Explainability (E), and Compliance (C).}
\resizebox{\textwidth}{!}{
\begin{tabular}{l|cccc}
\toprule
Data & Q & F & E & C \\
\midrule
\parbox{14cm}{A corpus of CO2 Electrocatalytic Reduction Process extracted from the scientific literature~\cite{mat-co2}} & 4.20 & 4.91 & 3.62 & 5.00 \\
\parbox{14cm}{Fatigue Database of Additively Manufactured Alloys~\cite{mat-fatigue}} & 4.10 & 4.38 & 4.00 & 4.00 \\
\parbox{14cm}{A materials terminology knowledge graph automatically constructed from text corpus~\cite{mat-term}} & 4.20 & 4.38 & 4.00 & 3.75 \\
\parbox{14cm}{Materials informatics platform with three dimensional structures (MIP-3d)~\cite{mat-mip3d}} & 4.29 & 4.25 & 3.75 & 4.00 \\
\parbox{14cm}{QM-symex-database~\cite{mat-qm}} & 4.30 & 4.25 & 3.50 & 4.00 \\
\parbox{14cm}{High Dielectric Ternary Oxides from Crystal Structure Prediction and High-throughput Screening~\cite{mat-crystal}} & 4.00 & 4.38 & 3.62 & 4.00 \\
\parbox{14cm}{QCDGE database: Quantum Chemistry Database with Ground- and Excited-State Properties~\cite{mat-quantum}} & 4.20 & 3.73 & 3.62 & 4.25 \\
\parbox{14cm}{Error assessment and optimal cross-validation approaches in machine learning applied to impurity diffusion~\cite{mat-error}} & 4.15 & 4.21 & 4.38 & 3.00 \\
\parbox{14cm}{Database of Open-Framework Aluminophosphate Structures~\cite{mat-al}} & 4.10 & 4.38 & 3.25 & 4.00 \\
\parbox{14cm}{A machine learning-based alloy design system to facilitate the rational design of high entropy alloys with enhanced hardness~\cite{mat-alloy}} & 4.15 & 4.21 & 4.31 & 3.00 \\
\bottomrule
\end{tabular}}
\label{tab:data_materials}
\end{table*}

\subsubsection{Prompt for Deductive Reasoning Assessment}
\label{app:deductive}
An example of a deductive reasoning task is shown below:

\begin{tcolorbox}[enhanced, colframe=black!75!white, colback=white, boxrule=0.3mm, sharp corners, left=1mm, right=1mm, top=1mm, bottom=1mm, fontupper=\footnotesize, before=\vspace{0.3em}, after=\vspace{1.3em}, width=\linewidth-2mm]
To identify a substance with an asymmetrical molecular structure, the following background knowledge is required: …\\
Using this foundation, analyze the molecular structures of the given substances to determine which one is asymmetrical.\\
\textbf{Question:} Which of the following substances has an asymmetrical molecular structure?\\
\textbf{Answer:}
\end{tcolorbox}


\subsection{Sub-dimensions of Values}
\label{app:subvalue}
We provide detailed evaluation dimensions for assessing whether LLMs align with the values that are essential in scientific work.

\begin{itemize}[leftmargin=20pt]
\item \textbf{Compliance with Laws and Regulations:} Ensuring that the decisions made are in accordance with relevant legal frameworks, such as data protection laws and intellectual property rights. This includes respecting the boundaries set by laws governing the research and application of AI.

\item \textbf{Academic Integrity:} Preventing unethical academic behaviors such as plagiarism, data manipulation, or fabrication. This involves making decisions that ensure transparency, honesty, and fairness in research, safeguarding the credibility of scientific work.

\item \textbf{Adherence to Objective Truth:} Prioritizing the use of accurate, unbiased data and scientific facts in decision-making. This value emphasizes the importance of avoiding distortion or misinformation in research findings, ensuring that only truthful and reliable data are used.

\item \textbf{Respect for Humanity and Nature:} Making decisions that respect human dignity, fundamental rights, and environmental sustainability. This value highlights the need to balance human welfare and ecological impact when making choices that affect society and the environment.

\item \textbf{Maintenance of Fairness and Justice:} Ensuring that decisions do not perpetuate biases related to gender, race, culture, or other demographic factors. Fairness and justice require that all individuals are treated equitably, and that decisions do not exacerbate existing societal inequalities.

\item \textbf{Attention to Privacy and Security:} Protecting personal data and ensuring that decisions respect privacy laws. This value underscores the importance of maintaining data security and preventing unauthorized access, breaches, or misuse of sensitive information.

\item \textbf{Ensuring Transparency and Explainability:} Making decisions that are transparent and explainable, with clear justifications that are accessible and understandable to users. Transparency and explainability ensure that the decision-making process is auditable and accountable, providing insight into how and why specific outcomes were reached.
\end{itemize}

\subsection{Expert Principles for Refining Questions of Value Assessment}
\label{app:valuefilter}

In the process of reviewing and filtering questions generated automatically, domain experts assess them based on several ethical principles as follows:

\begin{itemize}[leftmargin=20pt]
    \item \textbf{Relevance} – Ensuring realistic and discipline-specific scenarios.
    \item \textbf{Clarity} – Refining ambiguous or misleading phrasing.
    \item \textbf{Fairness} – Avoiding biased or culturally sensitive framing.
    \item \textbf{Validity} – Confirming that questions assess ethical principles rather than factual knowledge.
\end{itemize}

\section{More Details for Experimental Results}
\subsection{More Details for Scientific Data Assessment}
\label{app:data_bio}
We provide the results of data evaluation on Life and  Materials Sciences as shown in Table~\ref{tab:data_bio} and \ref{tab:data_materials}. 

\begin{table*}[htbp]
    \centering
\caption{Performance of LLMs on Knowledge assessment across multiple scientific disciplines.}
\resizebox{\textwidth}{!}{
\begin{tabular}{l|cccccc}
\toprule
Discipline&Overall&Life&Chemistry&Earth&Mathematics&Physics\\
\midrule
Gemini-2.5-pro-preview&\textbf{62.79}&\textbf{66.32}&\textbf{60.01}&\underline{64.29}&61.10&\textbf{62.20}\\
Gemini-2.5-flash-preview-thinking&\underline{60.09}&60.66&54.50&\textbf{66.95}&\underline{61.46}&\underline{56.87}\\
DeepSeek-R1&57.84&61.16&\underline{55.97}&58.80&59.80&53.48\\
Qwen3-235B-A22B&57.03&59.34&54.53&60.95&60.45&49.86\\
Gemini-2.0-Pro&56.83&\underline{61.49}&54.80&60.58&54.70&52.58\\
O3-Mini&54.90&56.41&54.33&56.78&57.46&49.51\\
Qwen2.5-Max&54.68&59.08&51.50&58.95&52.86&51.03\\
Claude-3.7-sonnet-thinking&54.30&60.00&46.65&58.78&56.18&49.89\\
DeepSeek-R1-Distill-Qwen-32B&54.09&60.18&45.96&50.81&\textbf{64.09}&49.41\\
Gemini-2.0-Flash&53.29&60.50&48.29&58.06&48.25&51.35\\
DeepSeek-V3&53.17&58.61&46.36&54.92&58.32&47.64\\
QwQ-32B&52.20&56.87&44.58&57.87&52.82&48.87\\
Claude-3.7-Sonnet&51.67&57.70&44.44&60.33&48.23&47.66\\
Qwen3-32B&51.59&52.29&47.63&57.00&54.56&46.44\\
Qwen2.5-72B-Instruct&51.01&57.87&47.91&53.51&51.16&44.62\\
Qwen-Plus&50.93&56.11&48.22&52.03&51.65&46.65\\
Doubao-Pro-32K&50.73&56.90&46.69&50.63&50.84&48.58\\
Claude-3.5-Sonnet-20241022&50.58&55.17&45.84&61.43&44.47&46.01\\
Grok-3&49.61&54.50&47.48&55.09&44.85&46.12\\
Gemini-2.5-flash-preview&49.58&53.59&42.49&57.27&48.46&46.07\\
InternLM2.5-20B-Chat&49.10&58.23&45.50&49.94&48.07&43.74\\
Qwen2-VL-7B-Instruct&48.01&55.76&48.85&52.97&39.16&43.30\\
Yi-34B-Chat&47.33&54.36&44.28&49.27&42.35&46.38\\
MiniCPM3-4B&47.32&54.08&46.52&48.27&43.86&43.89\\
ERNIE-4.0-Turbo-8K-Latest&47.25&54.60&42.10&47.42&46.81&45.34\\
GLM-4-Plus&46.92&53.68&46.80&50.25&38.96&44.91\\
Qwen3-30B-A3B&46.24&51.57&43.87&49.06&38.47&48.25\\
Llama3.1-70B-Instruct&46.23&56.02&44.77&45.17&43.09&42.13\\
GLM-4v-Plus&45.83&51.89&40.12&51.10&42.09&43.97\\
Llama3.3-70B-Instruct&45.49&54.32&45.51&45.33&40.34&41.94\\
O1-Mini-20240912&45.07&48.89&38.84&48.80&49.67&39.14\\
Llama-3.2-90B-Vision-Instruct&44.79&54.15&39.57&48.74&40.20&41.28\\
GPT-4.1&44.55&51.37&40.44&51.99&40.36&38.59\\
Llama-4-Maverick-17B-128E&44.32&52.11&43.85&46.07&37.15&42.44\\
Yi-Lightning&44.29&52.33&41.44&48.75&37.05&41.89\\
ERNIE-4.0-8K-Latest&44.22&50.23&45.25&47.73&39.75&38.15\\
Spark-v4.0-Ultra&44.18&50.22&47.32&46.39&37.48&39.51\\
GPT-4o-20241120&43.01&53.47&36.09&47.70&36.33&41.46\\
Mistral-Medium-3&42.94&49.26&35.74&48.03&41.38&40.28\\
InternLM3-8B-Instruct&42.33&50.97&43.97&44.05&35.12&37.57\\
Mistral-Small-24B-Instruct-2501&41.91&47.62&42.91&44.41&35.68&38.94\\
Moonshot-V1-32K&41.61&49.86&37.26&46.29&38.91&35.72\\
GLM-4-9B-Chat&41.46&45.37&41.28&42.73&38.54&39.37\\
Gemini-1.5-Pro-Latest&40.30&46.25&37.28&39.71&39.65&38.62\\
Qwen2.5-7B-Instruct&39.53&40.59&43.63&42.93&41.58&28.92\\
Yi-Vision-V2&39.43&46.32&40.57&41.15&35.05&34.08\\
Llama-4-Scout-17B-16E&39.30&44.54&36.61&41.36&37.83&36.16\\
MiniCPM-V-2.6&38.08&44.37&42.41&44.14&25.77&33.73\\
Ministral-8B-Instruct-2410&37.38&41.29&37.16&37.20&34.93&36.32\\
Pixtral-Large-Instruct-2411&36.92&43.97&33.11&40.60&32.44&34.46\\
Mistral-Large-Instruct-2411&35.98&43.13&36.59&39.04&28.99&32.13\\
GPT-4o-Mini&35.38&43.04&33.27&39.61&33.50&27.50\\
Llama-3.2-11B-Vision-Instruct&34.70&43.38&36.00&37.64&28.90&27.57\\
Llama3.1-8B-Instruct&33.69&37.90&34.63&37.24&28.10&30.57\\
Claude-3.5-Haiku-20241022&32.52&31.29&29.52&40.95&34.27&26.56\\
\bottomrule
\end{tabular}
    }
    \label{tab:knowledge}
\end{table*}

\begin{table*}[htbp]
    \centering
\caption{Performance of LLMs on Multimodality assessment across multiple scientific disciplines.}
\resizebox{\textwidth}{!}{
\begin{tabular}{l|cccccc}
\toprule
Discipline&Overall&Life&Chemistry&Earth&Mathematics&Physics\\
\midrule
Gemini-2.5-pro-preview&\textbf{79.85}&\textbf{90.69}&\textbf{76.63}&\underline{68.00}&\textbf{97.00}&\underline{66.94}\\
Gemini-2.0-Pro&\underline{68.03}&\underline{67.40}&\underline{73.22}&63.05&\underline{79.75}&56.72\\
Llama-4-Maverick-17B-128E&65.10&59.07&58.71&61.80&75.50&\textbf{70.43}\\
Claude-3.7-sonnet-thinking&62.31&52.94&67.76&67.86&70.50&52.51\\
Gemini-2.0-Flash&61.21&58.09&56.42&54.65&79.42&57.45\\
GPT-4.1&60.49&61.27&63.94&65.59&71.75&39.88\\
Claude-3.5-Sonnet-20241022&58.93&60.00&61.93&\textbf{69.55}&55.95&47.22\\
Gemini-1.5-Pro-Latest&56.88&60.05&48.87&65.33&64.25&45.88\\
GLM-4v-Plus&54.09&50.00&48.71&57.20&67.50&47.05\\
Claude-3.7-Sonnet&52.07&53.36&48.27&67.93&39.37&51.43\\
Qwen2.5-Max&51.63&60.30&50.06&52.66&48.16&46.95\\
Llama-4-Scout-17B-16E&51.51&48.77&45.93&60.44&46.25&56.14\\
Llama-3.2-90B-Vision-Instruct&51.40&57.60&54.47&55.03&56.50&33.43\\
GPT-4o-20241120&50.19&45.10&37.96&51.47&70.00&46.41\\
MiniCPM-V-2.6&48.72&48.04&42.80&44.55&59.75&48.44\\
Claude-3.5-Haiku-20241022&46.01&38.89&48.86&61.45&42.53&38.30\\
Qwen2-VL-7B-Instruct&40.76&34.07&50.95&48.20&33.75&36.84\\
Llama-3.2-11B-Vision-Instruct&40.72&46.08&38.69&36.95&47.75&34.12\\
GPT-4o-Mini&37.44&37.26&37.29&35.86&44.00&32.79\\
Yi-Vision-V2&35.89&19.61&31.41&37.39&62.75&28.32\\
\bottomrule
\end{tabular}
}
    \label{tab:multimodality}
\end{table*}

\subsection{More Details for LLM Assessment}
\label{app:llmevaluation}
We provide detailed experimental results that offer a comprehensive evaluation of LLMs across various scientific disciplines, as shown in Table \ref{tab:knowledge}, Table \ref{tab:understanding}, Table \ref{tab:reasoning}, Table \ref{tab:multimodality}, and Table \ref{tab:value}.

\begin{table*}[htbp]
    \centering
\caption{Performance of LLMs on Understanding assessment across multiple scientific disciplines.}
\resizebox{\textwidth}{!}{
\begin{tabular}{l|cccccc}
\toprule
Discipline&Overall&Life&Chemistry&Earth&Mathematics&Physics\\
\midrule
Gemini-2.5-pro-preview&\textbf{79.94}&\underline{65.52}&\textbf{100.00}&\underline{69.79}&\underline{91.75}&72.63\\
Gemini-2.5-flash-preview-thinking&\underline{78.28}&\textbf{66.44}&94.95&67.00&\textbf{92.00}&71.00\\
Gemini-2.0-Pro&77.60&60.27&\underline{98.00}&67.00&90.00&72.73\\
Llama-4-Maverick-17B-128E&77.01&65.07&94.00&67.00&85.00&\textbf{74.00}\\
Qwen3-235B-A22B&75.12&59.59&92.00&66.00&87.00&71.00\\
DeepSeek-R1&74.72&59.59&96.00&67.00&81.00&70.00\\
O3-Mini&74.08&64.38&97.00&63.00&79.00&67.00\\
GPT-4.1&73.75&64.38&93.00&63.64&73.74&\textbf{74.00}\\
DeepSeek-V3&73.48&64.38&85.00&\textbf{72.00}&82.00&64.00\\
Gemini-2.5-flash-preview&73.10&55.48&96.00&60.00&85.00&69.00\\
Claude-3.7-sonnet-thinking&71.19&51.37&97.00&59.60&78.00&70.00\\
Grok-3&70.51&57.53&96.00&61.00&74.00&64.00\\
GPT-4o-20241120&70.27&62.33&93.00&59.00&68.00&69.00\\
Gemini-2.0-Flash&69.22&52.74&93.94&58.00&75.76&65.66\\
Mistral-Medium-3&68.98&58.90&80.00&58.00&80.00&68.00\\
Claude-3.7-Sonnet&68.73&48.63&96.00&58.00&71.00&70.00\\
Qwen2.5-Max&67.92&54.79&83.00&59.00&79.80&63.00\\
Claude-3.5-Sonnet-20241022&66.92&46.58&95.00&55.00&67.00&71.00\\
Gemini-1.5-Pro-Latest&66.81&52.05&77.00&59.00&81.00&65.00\\
Llama-4-Scout-17B-16E&66.79&60.96&86.00&53.00&70.00&64.00\\
QwQ-32B&64.97&56.85&59.00&55.00&84.00&70.00\\
O1-Mini-20240912&64.64&58.22&65.00&49.00&86.00&65.00\\
Qwen3-32B&64.07&51.37&56.00&64.00&80.00&69.00\\
Qwen2.5-72B-Instruct&61.60&50.00&59.00&51.00&83.00&65.00\\
Yi-Lightning&61.20&50.00&66.00&55.00&68.00&67.00\\
Llama3.1-70B-Instruct&60.50&55.48&74.00&61.00&49.00&63.00\\
Qwen-Plus&60.33&48.63&59.00&57.00&78.00&59.00\\
Llama3.3-70B-Instruct&60.18&56.16&74.00&62.24&48.48&60.00\\
Pixtral-Large-Instruct-2411&58.95&39.73&63.00&54.00&67.00&71.00\\
GLM-4-Plus&56.82&41.10&64.00&50.00&67.00&62.00\\
DeepSeek-R1-Distill-Qwen-32B&56.72&42.86&63.39&54.00&67.36&56.00\\
ERNIE-4.0-Turbo-8K-Latest&54.96&41.78&63.00&48.00&63.00&59.00\\
Spark-v4.0-Ultra&54.71&44.60&53.19&54.17&63.27&58.33\\
Doubao-Pro-32K&54.38&45.89&58.00&42.00&75.00&51.00\\
Mistral-Small-24B-Instruct-2501&53.55&45.70&62.00&45.07&53.00&62.00\\
Claude-3.5-Haiku-20241022&53.16&30.82&74.00&48.00&54.00&59.00\\
Llama-3.2-90B-Vision-Instruct&51.00&41.78&62.24&53.00&40.00&58.00\\
GPT-4o-Mini&50.79&47.95&48.00&53.00&48.00&57.00\\
InternLM3-8B-Instruct&49.15&39.73&45.00&47.00&60.00&54.00\\
GLM-4v-Plus&49.01&39.04&42.00&42.00&53.00&69.00\\
Mistral-Large-Instruct-2411&47.55&39.73&47.00&38.00&48.00&65.00\\
Llama-3.2-11B-Vision-Instruct&47.06&43.15&50.75&47.00&42.42&52.00\\
InternLM2.5-20B-Chat&46.86&36.30&42.00&47.00&54.00&55.00\\
Moonshot-V1-32K&45.71&36.99&32.00&54.55&52.00&53.00\\
Qwen3-30B-A3B&44.59&23.97&53.00&45.00&75.00&26.00\\
Yi-34B-Chat&44.46&36.30&31.00&42.00&53.00&60.00\\
Llama3.1-8B-Instruct&43.78&45.89&39.00&46.00&38.00&50.00\\
MiniCPM3-4B&42.41&39.04&17.00&42.00&53.00&61.00\\
GLM-4-9B-Chat&41.72&35.62&25.00&44.00&46.00&58.00\\
ERNIE-4.0-8K-Latest&41.38&21.92&43.00&44.00&50.00&48.00\\
MiniCPM-V-2.6&41.21&39.04&24.00&45.00&45.00&53.00\\
Ministral-8B-Instruct-2410&40.89&42.47&31.00&43.00&41.00&47.00\\
Yi-Vision-V2&40.26&36.30&36.00&40.00&45.00&44.00\\
Qwen2-VL-7B-Instruct&38.19&34.93&15.00&36.00&55.00&50.00\\
Qwen2.5-7B-Instruct&32.76&30.82&5.00&32.00&52.00&44.00\\
\bottomrule
\end{tabular}
}
    \label{tab:understanding}
\end{table*}

\begin{table*}[htbp]
    \centering
\caption{Performance of LLMs on Reasoning assessment across multiple scientific disciplines.}
\resizebox{\textwidth}{!}{
\begin{tabular}{l|cccccc}
\toprule
Discipline&Overall&Life&Chemistry&Earth&Mathematics&Physics\\
\midrule
Gemini-2.5-pro-preview&\textbf{92.10}&\textbf{87.55}&\textbf{93.55}&96.23&\textbf{92.31}&90.84\\
Gemini-2.5-flash-preview-thinking&\underline{88.64}&82.78&\underline{83.41}&97.87&\underline{89.73}&89.42\\
O3-Mini&88.16&\underline{83.07}&80.68&\textbf{98.41}&87.44&\textbf{91.22}\\
Claude-3.7-sonnet-thinking&86.33&80.50&79.45&\underline{98.14}&82.58&\underline{91.00}\\
DeepSeek-R1&85.89&80.29&78.33&97.34&87.44&86.06\\
Qwen3-235B-A22B&85.35&82.73&79.16&97.61&82.03&85.24\\
O1-Mini-20240912&85.25&80.50&77.65&95.97&87.09&85.05\\
Gemini-2.0-Pro&82.40&81.64&76.74&\underline{98.14}&73.65&81.84\\
QwQ-32B&82.32&76.28&76.74&97.08&83.42&78.09\\
Llama-4-Maverick-17B-128E&81.86&81.36&79.85&93.25&72.95&81.87\\
Qwen3-32B&80.63&75.43&71.21&\underline{98.14}&81.69&76.69\\
Qwen2.5-Max&80.46&77.93&78.86&92.73&69.45&83.36\\
Llama-4-Scout-17B-16E&80.14&76.28&74.92&90.29&78.97&80.25\\
DeepSeek-V3&79.66&78.86&79.62&92.46&64.92&82.45\\
Gemini-1.5-Pro-Latest&79.07&77.72&82.88&92.99&62.68&79.07\\
Gemini-2.5-flash-preview&78.64&81.64&77.95&94.64&64.27&74.69\\
Mistral-Medium-3&76.39&73.43&73.71&93.25&63.73&77.85\\
Claude-3.7-Sonnet&76.27&74.07&74.92&94.90&54.44&83.02\\
Gemini-2.0-Flash&75.10&75.19&69.97&94.90&59.05&76.40\\
Claude-3.5-Sonnet-20241022&74.89&74.65&78.33&95.17&53.20&73.09\\
GPT-4.1&74.39&82.33&77.72&97.86&43.67&70.38\\
Grok-3&73.77&72.93&74.62&91.92&58.87&70.49\\
GPT-4o-20241120&72.36&80.78&73.79&97.08&41.05&69.11\\
Qwen-Plus&72.18&71.71&71.89&87.58&55.89&73.84\\
Qwen2.5-72B-Instruct&72.13&70.07&72.80&91.66&53.25&72.87\\
Llama3.3-70B-Instruct&71.45&77.93&72.19&90.29&46.41&70.42\\
Llama3.1-70B-Instruct&71.06&76.00&71.59&90.29&46.95&70.49\\
Doubao-Pro-32K&70.81&67.28&75.15&87.63&51.72&72.27\\
DeepSeek-R1-Distill-Qwen-32B&69.31&70.93&63.75&74.14&69.38&68.34\\
GLM-4-Plus&69.24&68.93&71.29&92.77&45.91&67.29\\
Llama-3.2-90B-Vision-Instruct&69.12&69.77&70.61&90.56&51.03&63.63\\
Spark-v4.0-Ultra&67.32&73.32&67.46&86.61&49.04&60.16\\
ERNIE-4.0-Turbo-8K-Latest&67.01&65.85&69.85&91.71&41.54&66.09\\
Mistral-Small-24B-Instruct-2501&66.84&68.81&69.27&88.02&45.27&62.86\\
Mistral-Large-Instruct-2411&66.66&67.43&70.38&93.84&38.96&62.69\\
Yi-Lightning&66.23&70.00&63.56&89.49&45.96&62.13\\
Moonshot-V1-32K&65.77&72.72&70.68&80.31&41.05&64.09\\
GLM-4v-Plus&65.74&66.86&72.58&77.61&39.46&72.22\\
GPT-4o-Mini&65.19&72.86&66.06&86.25&35.09&65.71\\
Pixtral-Large-Instruct-2411&65.07&70.38&69.47&94.10&35.44&55.95\\
InternLM2.5-20B-Chat&64.93&61.28&66.37&80.84&52.27&63.89\\
InternLM3-8B-Instruct&64.79&72.78&71.06&77.55&41.05&61.51\\
Claude-3.5-Haiku-20241022&64.46&71.35&69.55&86.30&37.38&57.73\\
Llama-3.2-11B-Vision-Instruct&62.58&61.95&64.08&84.17&43.75&58.98\\
Yi-34B-Chat&61.94&65.22&60.83&92.24&27.95&63.49\\
GLM-4-9B-Chat&61.17&61.57&67.65&73.78&43.13&59.71\\
Llama3.1-8B-Instruct&60.61&58.78&65.76&81.73&43.47&53.31\\
MiniCPM3-4B&60.44&60.71&60.53&77.07&44.37&59.51\\
Ministral-8B-Instruct-2410&60.31&63.50&69.47&79.29&31.96&57.31\\
Yi-Vision-V2&58.81&56.00&61.44&78.18&43.33&55.11\\
ERNIE-4.0-8K-Latest&57.75&55.65&59.85&81.73&33.01&58.51\\
Qwen2.5-7B-Instruct&57.66&53.64&67.95&81.95&31.62&53.11\\
Qwen2-VL-7B-Instruct&55.49&49.79&63.94&77.61&27.40&58.71\\
MiniCPM-V-2.6&54.52&53.22&60.53&67.89&32.45&58.54\\
Qwen3-30B-A3B&53.60&48.14&62.04&76.81&51.86&29.15\\
\bottomrule
\end{tabular}
}
    \label{tab:reasoning}
\end{table*}

\begin{table*}[htbp]
    \centering
\caption{Performance of LLMs on Values assessment across multiple scientific disciplines.}
\resizebox{\textwidth}{!}{
\begin{tabular}{l|cccccc}
\toprule
Discipline&Overall&Life&Chemistry&Earth&Mathematics&Physics\\
\midrule
Claude-3.7-sonnet-thinking&\textbf{69.90}&\underline{65.99}&\underline{72.53}&70.29&\textbf{71.05}&\textbf{69.67}\\
DeepSeek-R1&\underline{68.27}&61.39&69.54&\textbf{78.46}&63.16&\underline{68.81}\\
Claude-3.5-Sonnet-20241022&67.67&64.12&\textbf{73.04}&74.46&63.75&62.99\\
Claude-3.7-Sonnet&66.65&61.90&69.32&71.50&\underline{67.17}&63.37\\
Gemini-1.5-Pro-Latest&66.33&\textbf{66.07}&68.64&75.04&63.85&58.03\\
Mistral-Small-24B-Instruct-2501&66.14&63.01&64.63&\underline{76.96}&62.65&63.45\\
Claude-3.5-Haiku-20241022&66.03&62.33&66.83&69.52&63.68&67.78\\
O3-Mini&64.90&62.67&66.19&67.63&64.02&64.00\\
GLM-4v-Plus&64.03&61.48&64.56&66.01&64.54&63.57\\
O1-Mini-20240912&63.36&55.95&60.36&71.44&64.02&65.04\\
Qwen3-235B-A22B&62.98&63.01&63.99&71.64&58.93&57.32\\
DeepSeek-V3&62.86&60.89&60.50&75.02&61.06&56.83\\
DeepSeek-R1-Distill-Qwen-32B&62.77&59.01&63.31&68.80&61.89&60.83\\
GPT-4o-20241120&62.31&63.27&61.70&70.07&61.92&54.59\\
InternLM3-8B-Instruct&62.03&59.35&59.54&68.59&58.42&64.26\\
Gemini-2.5-flash-preview&62.01&60.21&61.69&68.35&60.69&59.12\\
Gemini-2.5-flash-preview-thinking&61.80&60.54&63.64&70.10&60.85&53.88\\
Llama-3.2-90B-Vision-Instruct&61.71&58.33&64.85&75.02&63.48&46.85\\
Spark-v4.0-Ultra&61.65&59.88&63.88&65.20&59.45&59.84\\
Qwen3-32B&60.85&60.03&61.84&69.63&61.51&51.21\\
Yi-Lightning&60.74&55.02&61.58&70.37&61.62&55.10\\
Gemini-2.0-Flash&60.33&63.52&60.43&69.58&57.86&50.28\\
Yi-34B-Chat&60.14&57.82&59.66&65.98&59.12&58.13\\
GLM-4-9B-Chat&59.37&56.63&57.68&68.87&56.78&56.91\\
Mistral-Large-Instruct-2411&59.27&58.84&55.38&69.71&60.70&51.73\\
Mistral-Medium-3&59.25&55.95&58.78&73.30&57.83&50.41\\
Llama3.1-70B-Instruct&59.00&57.57&56.24&74.47&60.48&46.24\\
Llama3.3-70B-Instruct&58.91&57.23&56.24&73.90&60.48&46.69\\
Qwen3-30B-A3B&58.49&59.35&61.84&66.66&51.40&53.20\\
Llama-3.2-11B-Vision-Instruct&58.29&57.40&55.00&67.57&55.38&56.10\\
Llama-4-Scout-17B-16E&58.08&50.77&61.68&63.47&53.98&60.49\\
GLM-4-Plus&58.06&56.38&58.05&64.92&63.60&47.35\\
GPT-4o-Mini&57.58&52.21&51.07&70.70&63.86&50.08\\
Llama-4-Maverick-17B-128E&57.31&47.70&62.56&64.02&52.44&59.81\\
InternLM2.5-20B-Chat&57.22&49.40&58.35&67.33&48.19&62.83\\
Yi-Vision-V2&57.12&53.23&61.37&66.99&54.13&49.86\\
Grok-3&57.06&49.24&57.48&64.98&52.17&61.44\\
Gemini-2.5-pro-preview&56.90&57.79&55.26&66.38&59.37&45.70\\
Llama3.1-8B-Instruct&56.88&55.78&54.40&65.67&53.25&55.29\\
Qwen2-VL-7B-Instruct&56.14&52.12&58.29&63.70&46.75&59.82\\
Moonshot-V1-32K&56.09&60.12&53.18&65.64&51.59&49.95\\
Qwen-Plus&55.90&60.71&45.80&68.45&56.81&47.71\\
ERNIE-4.0-Turbo-8K-Latest&55.53&51.44&54.83&67.41&57.82&46.14\\
Qwen2.5-Max&55.42&57.99&46.85&62.44&62.25&47.59\\
MiniCPM3-4B&55.38&55.87&52.96&60.62&53.87&53.60\\
Gemini-2.0-Pro&55.23&55.53&51.65&69.46&53.39&46.13\\
Pixtral-Large-Instruct-2411&55.07&55.19&52.84&64.84&56.48&46.01\\
Doubao-Pro-32K&54.86&57.06&54.99&68.13&49.78&44.35\\
GPT-4.1&54.60&56.98&56.55&61.00&54.70&43.77\\
QwQ-32B&54.07&51.53&45.54&65.38&50.26&57.62\\
Qwen2.5-7B-Instruct&53.08&55.61&48.33&66.04&45.47&49.95\\
MiniCPM-V-2.6&52.31&45.49&59.42&59.17&44.24&53.24\\
Qwen2.5-72B-Instruct&49.55&50.00&41.38&61.80&50.70&43.88\\
Ministral-8B-Instruct-2410&46.79&46.17&48.74&51.31&41.15&46.59\\
ERNIE-4.0-8K-Latest&41.57&37.84&43.98&51.18&40.55&34.31\\
\bottomrule
\end{tabular}
}
    \label{tab:value}
\end{table*}
 \clearpage
\bibliography{reference,ref2}

\end{document}